\pdfoutput=1

\documentclass[11pt]{article}

\usepackage{ACL2023}

\usepackage{times}
\usepackage{hyperref}
\usepackage{latexsym}
\usepackage{amsmath,amsfonts,amssymb}
\usepackage[T1]{fontenc}

\usepackage[utf8]{inputenc}
\usepackage{graphicx}
\usepackage{svg}

\usepackage{microtype}
\usepackage{amsfonts,amsmath,amssymb,amsthm}
\usepackage{inconsolata}
\usepackage{cleveref}
\usepackage{paralist}
\usepackage[shortlabels]{enumitem}
\usepackage{xurl}
\setlist{topsep=2pt, leftmargin=*, itemsep=1pt}
\title{On the Correspondence between Compositionality and Imitation in Emergent Neural Communication}

\author{Emily Cheng\thanks{~ Work done while visiting LATTICE at the CNRS / École Normale Supérieure.} \\
  UPF, Barcelona \\
  \texttt{emilyshana.cheng@upf.edu} \\\And
  Mathieu Rita \\
  INRIA, Paris \\
  \texttt{mathieu.rita@inria.fr} \\\And
  Thierry Poibeau \\
  CNRS \& ENS-PSL, Paris  \\
  \texttt{thierry.poibeau@ens.psl.eu}}

\begin{document}
\maketitle
\begin{abstract}
Compositionality is a hallmark of human language that not only enables linguistic generalization, but also potentially facilitates acquisition. When simulating language emergence with neural networks, compositionality has been shown to improve communication performance; however, its impact on imitation learning has yet to be investigated. Our work explores the link between compositionality and imitation in a Lewis game played by deep neural agents. Our contributions are twofold: first, we show that the learning algorithm used to imitate is crucial: supervised learning tends to produce more average languages, while reinforcement learning introduces a selection pressure toward more compositional languages. 
Second, our study reveals that compositional languages are easier to imitate, which may induce the pressure toward compositional languages in RL imitation settings.
\end{abstract}

\section{Introduction}


Compositionality, a key feature of human language, makes it possible to derive the meaning of a complex expression from the combination of its constituents~\citep{sep-compositionality}. It has been suggested that more compositional languages are easier to acquire for both humans and artificial agents~\cite{RAVIV2021104620,liandbowling2019,Ren2020Compositional,chaabouni-etal-2020-compositionality}. Therefore, to better understand the factors underlying language transmission, it is crucial to understand the relationship between ease-of-acquisition and compositionality.

We study the link between compositionality and ease-of-acquisition in the context of emergent communication. In this setting, two deep artificial agents with asymmetric information, a Sender and a Receiver, must develop communication from scratch in order to succeed at a cooperative game~\citep{havrylov_titov2017,lazaridou2017multiagent,lazaridou2020emergent}. We will refer to this mode of language learning, in which agents develop language via feedback from mutual interaction, as \emph{communication-based learning}. 

Several studies have linked compositionality to ease-of-acquisition in communication-based learning.~\citealp{chaabouni-etal-2020-compositionality} show compositionality predicts efficient linguistic transmission from Senders to new Receivers. Conversely,~\citealp{liandbowling2019} re-pair a Sender periodically with new Receivers, and show this ease-of-teaching pressure improves compositionality. 


Communication-based learning is not the only possibility for language learning, however. Humans also crucially acquire language through \emph{imitation-based learning}, in which they learn by observing other humans' language use~\citep{Kymissis_Poulson_1990}.~\citealp{Ren2020Compositional} and~\citealp{chaabouni-etal-2020-compositionality} employ imitation learning, where in the first study, agents undergo a supervised imitation stage before communication-based learning, and where in the second, agents alternate between communication-based learning and imitating the best Sender. However, the dynamics of imitation are not the focus in either study.  For such an important vehicle of language acquisition, imitation-based learning thus remains under-explored in the emergent communication literature. 

 We extend these lines of inquiry to systematically investigate compositionality in imitation-based learning.\footnote{...for artificial agents. We do not test theories of human imitation learning.} Our contributions are as follows: (1) We show that imitation can automatically select for compositional languages under a reinforcement learning objective; and (2) that this is likely due to ease-of-learning of compositional languages. 


\section{Setup}
We study imitation in the context of referential communication games~\citep{lewis1969convention}. In this setting, a Sender agent observes an object $x$ and transmits a message $m$ to a second Receiver agent. Using this message, the Receiver performs an action for which both agents receive a reward. Over the course of the game, agents converge to a referential system $(x,m)$, which we refer to as an emergent language. 

\paragraph{Measuring Compositionality}
Evaluating compositionality in emergent languages is not straightforward given their grammars are a-priori unknown. Therefore, we quantify compositionality using topographic similarity (topsim)~\cite{kirby2006}, a grammar-agnostic metric widely applied to emergent languages in the literature. Topsim is defined as the Spearman correlation $\rho$ between Euclidean distances in the input space and Levenstein distances in the message space-- that is, it captures the intuition that nearby inputs should be described with similar messages. While we consider other compositionality metrics such as positional disentanglement~\cite{chaabouni-etal-2020-compositionality}, we focus on topsim due to its high correlation with generalization accuracy ($\rho = 0.83$)~\citep{rita2022emergent}. See \cref{sec:other_compo} for extended experiments on compositionality metrics and generalization.

\subsection{Imitation Task}
\begin{figure}
    \centering
    \includegraphics[width=\linewidth]{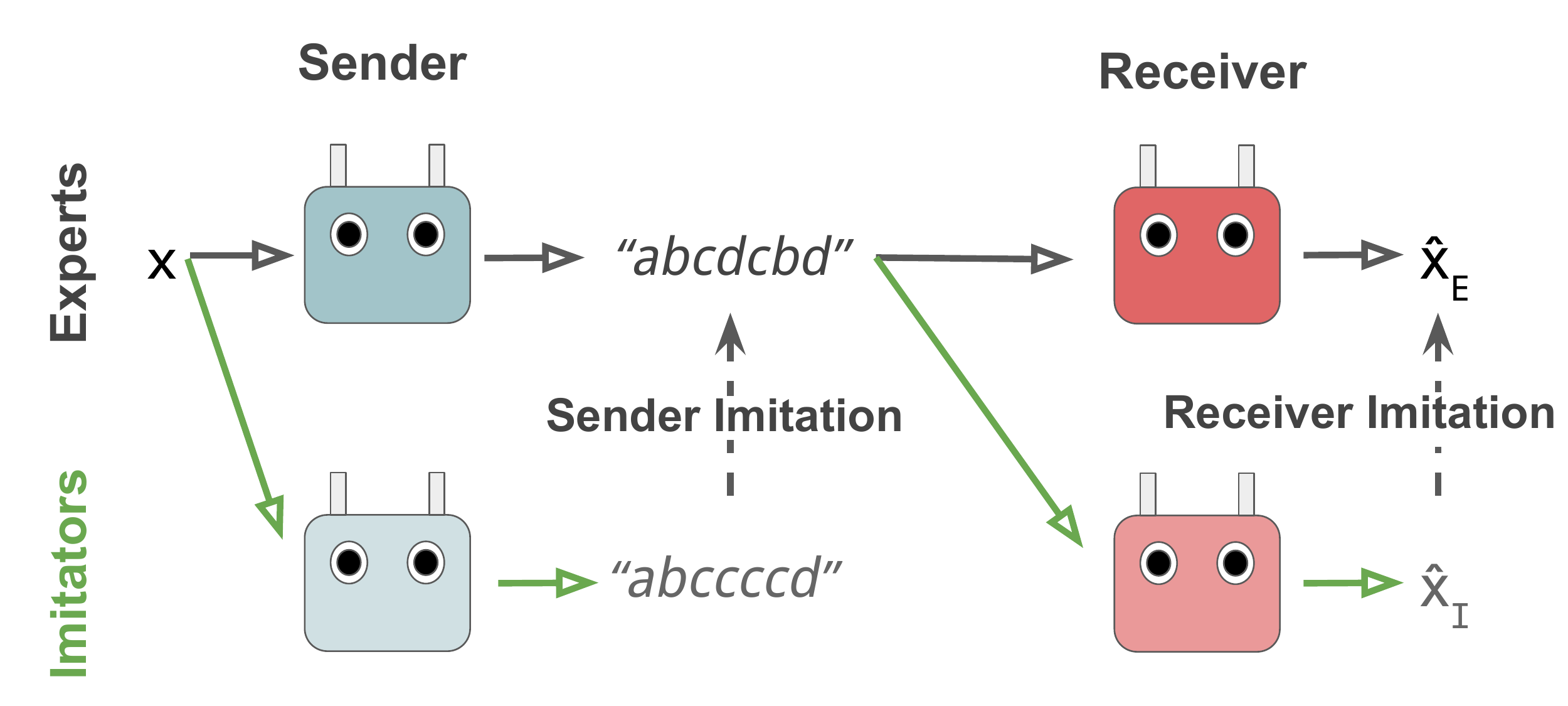}
    \caption{Imitation setup for one Expert pair of agents. A newly initialized Imitator Sender (lower left) and Imitator Receiver (lower right) imitate an Expert Sender (top left) and an Expert Receiver (top right). The Expert Sender has been trained on a communication game with the Expert Receiver (bold arrows), so that when the Sender encodes input $x$ into message $m$ (e.g., ``abcdcbd"), the Receiver decodes $m$ into $\hat x_E$, reconstructing $x$. Imitation (dotted arrows) consists of training the Imitator on inputs/outputs of the respective Expert: ($x$, $m$) for the Sender and ($m$, $\hat x_E$) for the Receiver.}
    \label{fig:setup}
\end{figure}

To investigate whether compositional languages are selected for in imitation, we posit an imitation task where one new \emph{Imitator} Sender or Receiver simultaneously imitates several \emph{Expert} Senders or Receivers with varying topsims. Both Sender and Receiver agents are parameterized by single-layer GRUs~\cite{cho-etal-2014-properties} that are deterministic after training (see \cref{sec:implementation} for implementation).\footnote{Experiments are implemented using EGG~\cite{kharitonov:etal:2021}. Code may be found at \\ \small{\url{https://github.com/chengemily/EGG/tree/imitation}}.} While we explore imitation for both agents, we focus on Sender imitation in the main text, and extend to Receiver imitation in \cref{sec:receiver_imitation}. A minimal example of imitation learning with only one Expert Sender-Receiver pair is shown in \cref{fig:setup}.

The Sender imitation task is as follows: given a set of $k$ Expert Senders, we train an identical, newly initialized Sender on the Experts' inputs and outputs $(x,m)$. That is, for each round of training, all $k$ Experts as well as the Imitator Sender receive input $x$ and output $m^{(1)} \cdots m^{(k)}$ and $m^I$, respectively. The Imitator is then tasked to minimize the difference between their output and a uniform mixture of the $k$ Expert outputs.

\paragraph{Dataset} Data in the imitation task consists of inputs and outputs of pairs of Expert agents trained to convergence on a communication game-- in our case, the two-agent reconstruction task of~\citet{kottur2017natural}. To generate the Experts, we pre-train $N=30$ Sender-Receiver pairs on this reconstruction task to high validation accuracy ($0.99 \pm 0.01$) (task and training details in \cref{sec:expert_pretrain}).

Expert training produces the following data: 1) inputs $x$; 2) messages $m$ corresponding to Expert Senders' encodings of $x$; and 3) outputs $\hat x$, the Expert Receivers' reconstructions of $x$ given $m$.

Each input $x$ denotes an object in an ``attribute-value world", where the object has $n_{att}$ attributes, and each attribute takes $n_{val}$ possible values. We represent $x$ by a concatenation of $n_{att}$ one-hot vectors, each of dimension $n_{val}$. On the other hand, messages $m$ are discrete sequences of fixed length $L$, consisting of symbols taken from a vocabulary $V$. We set $n_{att} = 6$, $n_{val}=10$, $|V| = 10$, and $L=10$, corresponding to a relatively large attribute-value setting in the literature (Table 1 of~\citet{raviv_directions}). 

We split the input data ($n = 10^6$) into a training set and two holdout sets. Similar to~\citet{chaabouni-etal-2020-compositionality}, we define two types of holdouts: a zero-shot generalization set ($n=354294$), where one value is held-out during training, and an in-distribution generalization set ($n=531441$). The training set, on which we both train and validate, represents 1\% of in-distribution data ($n=5315$). These data splits are used in Expert training and are inherited by the imitation task (see \cref{sec:hyperparams} for details on generating the data split).




\paragraph{Imitation learning algorithms} While imitation is classically implemented as supervised learning, we test two imitation learning procedures: 1) supervised learning (SV) with respect to the cross-entropy loss between Imitator and Expert outputs; and 2) reinforcement learning with the REINFORCE algorithm (RF)~\cite{reinforce}, using per-symbol accuracy as a reward. When using REINFORCE, we additionally include an entropy regularization term weighted by $\lambda$ to encourage exploration, and subtract a running mean baseline from the reward to improve training stability~\citep{Williams_Peng_1991}. See \cref{sec:app_selection} for loss functions and \ref{sec:hyperparams} for detailed hyperparameter settings.


\subsection{Evaluation}
To evaluate properties of imitation learning, we identify three descriptors of interest: validation accuracy, ease-of-imitation, and selection of compositional languages. 

\paragraph{Accuracy} We evaluate imitation performance between an Imitator and Expert by the average per-symbol accuracy between their messages given an input. When using REINFORCE, training accuracy is computed using the Imitators' sampled output while validation accuracy is computed using the Imitators' argmax-ed output.

\paragraph{Ease-of-imitation} We evaluate ease-of-imitation of a language two ways. First, imitation sample complexity ($T$): the number of epochs needed to reach 99\% validation accuracy, and second, imitation speed-of-learning ($\texttt{SOL}^I$): the area under the validation accuracy curve, cut-off at $t$ epochs chosen by visual inspection of convergence.

\paragraph{Selection of compositional languages}
Sender imitation consists of learning one-to-one input-to-message mappings from a sea of one-to-many Expert mappings.
Then, the Imitator's language will consist of a mixture of Expert languages, where the mixture weights reveal the extent of selection.

In this mixture, we proxy the Imitator's learned weight for an Expert as the proportion of messages in the training set for which Imitator accuracy on the Expert message is the highest.  Note that the coefficients may not add to one: if the highest Expert accuracy for a message does not exceed chance (10\%), we consider the message unmatched. 

To quantify selection, we use the intuition that selection corresponds jointly to peakedness and asymmetry in the learned distribution over Expert languages sorted by topsim. We evaluate peakedness using the Shannon entropy and asymmetry using Fisher's moment coefficient of skew of Expert weights. Formally, let there be $k$ Experts, where Experts are sorted in ascending order of topsim (Expert $i$=1 is the least and $i$=$k$ is the most compositional, respectively). The Imitator learns a mixture of the Expert languages with weights $W:=(w_{i})_{1 \leq i \leq k}$ 
(normalized). Given $W$, we evaluate peakedness with:
\begin{align}
    \mathcal H(W) = - \sum_{i=1}^k w_i \log (w_i).
\end{align}
To quantify asymmetry of expert weights, we estimate the Fisher's moment coefficient of skew:
\begin{align}
\tilde{\mu}(W) &= \frac{1}{k} \sum_{i=1}^k \left(\frac{w_i - \mu}{\sigma}\right)^3,
\end{align}

where $\mu$ is the mean and $\sigma$ is the standard deviation of $W$. A skew of 0 implies perfect symmetry, positive skew corresponds to a right-tailed distribution, and negative skew corresponds to a left-tailed distribution. Intuitively, the more negative the skew of the Expert weights, the more weight lies on the right side of the distribution, hence the greater ``compositional selection effect". 

We proxy selection, then, by a negative skew (more weight assigned to high-topsim Experts) and low entropy (peakedness) in the Expert weight distribution.

\section{Imitation and Selection of Compositionality}
We present results for imitation on mixtures of $k=2$-$5$ Expert Senders. First, we generate 30 Expert languages from the referential task, initially considering Expert distributions corresponding to evenly-spaced percentiles of topsim, including the minimum and maximum $(0.26, 0.43)$. For example, when $k=3$, we take the lowest, $50^{th}$ percentile, and highest-topsim languages. All results are aggregated over $5$ random seeds after $2000$ training epochs.

We find that (1) whether Imitators prefer compositional Experts depends crucially on the learning algorithm: imitation by reinforcement results in marked compositional selection compared to supervision; and (2) compositional selection also depends on variance of expert topsims, $\lambda$ entropy regularization coefficient, and number of Experts.
 
 \begin{figure}[h]
\includegraphics[width=0.44\textwidth]{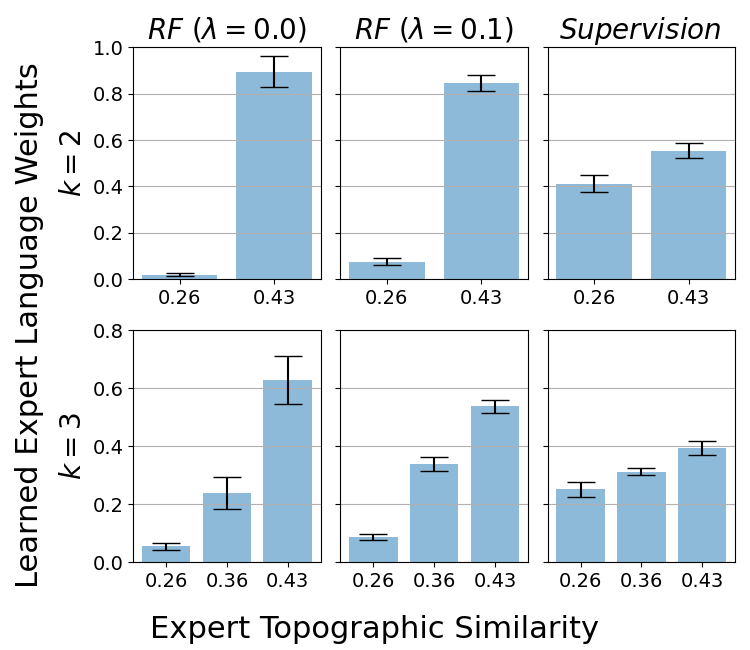}
\caption{Sender Imitator's learned weights ($\pm 1$ std.) on $k=2$ (top) and $k=3$ (bottom) Expert languages whose topsims range evenly from 0.26 to 0.43. The left two columns correspond to imitation by reinforcement (RF). As the entropy coefficient $\lambda$ increases (left to middle), the weights are more uniform, and are most uniform in the supervised setting (right). Refer to \cref{fig:skew_entropy_2_3} for skews and entropies of the distributions.}
 \label{fig:sender_comp}
\end{figure}

 \begin{figure}[h!]
 \centering
\includegraphics[width=0.47\textwidth]{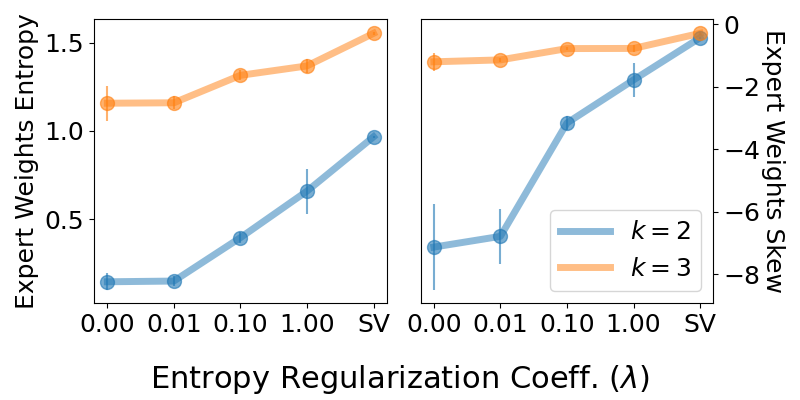}
\caption{Entropy (left) and skew (right) ($\pm 1$ std.) of learned Expert weights by a Sender Imitator for $k=2$ and $3$ Experts. Expert languages' topsims range evenly from 0.26 to 0.43. Both entropy and skew increase to the entropy of a uniform distribution, skew of a symmetric distribution ($=0$), respectively as exploration ($\lambda$) increases, attaining maxima in supervision (SV).}
\label{fig:skew_entropy_2_3}
 \end{figure}




The distribution of learned Expert weights in \cref{fig:sender_comp}, as well as imitation validation accuracy curves in \cref{fig:imi_val}, evidence that in imitation by supervision, the empirical mixture is closer to uniform than when imitating by reinforcement. Otherwise, when optimizing using reinforcement, the Imitator selects more compositional languages. 

The shape of the Expert weight distribution is tempered by the entropy regularization coefficient $\lambda$: smaller $\lambda$ results in greater compositional selection (that is, lower entropy and more negative skew) of the weight distribution (\cref{fig:skew_entropy_2_3}). At the limit, imitation by supervision results in the highest entropy and the skew that is closest to zero. 

\begin{figure}[h!]
\centering
\includegraphics[width=0.47\textwidth]{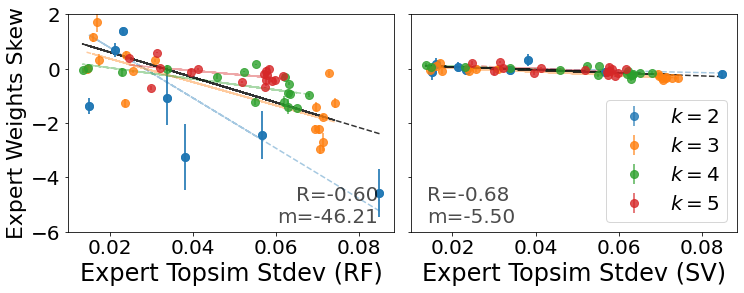}
\caption{The skew of learned Expert Sender weights vs. the standard deviation of the Expert topsim ($\pm 1$ std.) for RF (left) and SV (right) for $k=2$-$5$ Experts. Expert weight skew and Expert topsim standard deviation are highly and significantly correlated ($\alpha=1$e-$6$), and the linear effect $m$ is much (6x) higher for RF than for SV.}
\label{fig:skew_topsimentropy_all}
 \end{figure}

We then test the effect of Expert topsim distribution \emph{asymmetry} on the learned weights. To do so, for each $k>2$, we generate $10$ Expert topsim distributions with varying skew, following the procedure outlined in \cref{sec:shape} (when $k=2$, skew is mechanically $0$). We find that for both REINFORCE and supervision, holding $k$ equal, the skew and entropy of the learned Expert weight distribution are robust (i.e., not correlated) to the skew of the underlying Expert topsim distribution (\cref{fig:robust_to_skew_entropy_composite}). This is desirable when imitating by reinforcement and undesirable when imitating by supervision: for example, consider Expert topsim distributions [\texttt{low} \texttt{high} \texttt{high}] (skew$<0$) and [\texttt{low} \texttt{low} \texttt{high}] (skew$>0$). In both cases, REINFORCE will select a high-topsim Expert, and supervision will weight all Experts equally, that is, supervision is unable to de-select poor topsims. 

Using all Expert topsim distributions generated so far (those where topsim ranks are evenly spaced, and those exhibiting varying skews), we investigate the effect of topsim distribution \emph{spread}, quantified by standard deviation, on the learned weights. In \cref{fig:skew_topsimentropy_all}, we note a significant negative effect of Expert topsim standard deviation on the degree of compositional selection. That is, the more disperse the Expert topsims, the more the Imitator can differentiate between and select compositional Experts (shown by a more negative skew in learned Expert weights). Though this correlation is highly statistically significant for both REINFORCE and supervision, the effect is $\sim8$x greater for REINFORCE, demonstrating that the spread between expert compositionalities plays a more important role in the degree of selection by reinforcement. 

Finally, selection is less salient as the number of Experts increases, seen by the increasing entropies and skews of Expert weights (\cref{fig:skew_entropy_2_3,fig:skew_entropy_4_5}). Results for $k>3$ may be found in \cref{sec:app_selection}.

 \paragraph{Understanding why REINFORCE selects for compositional languages}
 The different results between the optimization algorithms correspond to inherent differences in learning objective. Successful imitation minimizes the Kullback-Leibler divergence between the Imitator $\pi^I$ and the Expert policies $\pi^E$; supervision is classically known to minimize the \emph{forward} KL divergence $D_{KL}(\pi^E||\pi^I)$, while reinforcement minimizes the \emph{reverse} KL divergence $D_{KL}(\pi^I||\pi^E)$ with respect to $\pi^I$. That is, imitation by supervision is mean-fitting while imitation by reinforcement is mode-fitting-- the former learns a uniform mixture of Expert languages (see \cref{sec:appendix_proof} for proof), and the latter selects the best Expert language. 


\section{Speed-of-Imitation May Explain Compositional Selection}
\label{sec:sol}
Thus far, we have seen that imitation by reinforcement selects compositional languages. This is likely because higher topsim languages are \emph{easier to imitate}. We establish a positive and statistically significant relationship between topsim and ease-of-imitation, expanding the explorations in~\citet{Ren2020Compositional,liandbowling2019,chaabouni-etal-2020-compositionality} (see \cref{sec:appendix} for experimental details).
 
 We evaluate ease-of-imitation using $k=1$, after $t=500$ (SV) and $2000$ epochs (RF), where $t$ is chosen based on validation accuracy convergence. Correlations between topsims of $30$ Expert languages and Imitator performance (averaged over three random seeds) are shown in \cref{tab:correlations}. We find that for both imitation by supervision and reinforcement, topsim is (1) significantly negatively correlated to imitation sample complexity $T$; (2) significantly positively correlated to speed-of-imitation \texttt{SOL}. 



\begin{table}[h]
    \centering
    \scalebox{0.9}{
    \begin{tabular}{rl|c c c c }
    & & $T_S$ & $T_R$ & \texttt{SOL}$_S^I$ & \texttt{SOL}$_R^I$ \\
    \hline 
    \textbf{SV} & $\rho$ & -0.65 & -0.80 & 0.65 & 0.75\\
      & $R^2$ & -0.66 & -0.80 & 0.65 & 0.76 \\ 
      \hline
    \textbf{RF} & $\rho$ & -0.66 & -0.60  & 0.45 & 0.59 \\
        & $R^2$ & -0.66 & -0.68 & 0.41* & 0.63 \\
        \hline
    \end{tabular}}
    \caption{Spearman $\rho$ and Pearson's $R$ between Expert topsim and Imitator learning speed (Sender=$S$, Receiver=$R$). Unless otherwise stated, all correlations are significant using $\alpha = 1$e-$2$. *($\alpha=0.05$)}
    \label{tab:correlations}
\end{table}

Moreover, correlations between topsim and ease-of-imitation are stronger than those between Expert validation accuracy and ease-of-imitation (\cref{tab:spearman_bc}). This suggests that the positive relationship between compositionality and ease-of-imitation is not due to a confound of high validation accuracy. 

 \section{Discussion}
Having (1) demonstrated a selection of compositional languages in imitation by reinforcement; (2) established a significant correlation between topsim and ease-of-imitation; we offer the following explanation for compositional selection: \emph{mode-seeking behavior in reinforcement learning exploits ease-of-learning of compositional languages, resulting in a selection of compositionality.}

While both imitation and ease-of-learning of compositional languages have been instrumentalized in population training, they are engineered in a top-down way: in~\citet{chaabouni2022emergent}, agents imitate the best-accuracy agent, who is algorithmically designated as the teacher; in~\citet{Ren2020Compositional}, imitation is stopped early to temporally select compositional features.\footnote{We did not succeed in replicating results in~\citet{Ren2020Compositional} (see \cref{sec:appendix}).} Our work, using basic RL principles, proposes an alternative mechanism that selects compositional languages while requiring minimal engineering and assumptions.

Selection by RL imitation, using the same ease-of-learning argument, applies to not only compositionality but also potentially to other traits, e.g., language entropy or message length.  That is, RL imitation \emph{naturally promotes any learnability advantage} among candidate languages without manual intervention, while \emph{agnostic to the signaling system}. This may then be leveraged alongside communication-based learning in population-based emergent communication, where imitation would persist easy-to-learn linguistic features.

\section*{Limitations}

There are several limitations to our work. 

First, although we choose the attribute-value dataset due to its high degree of interpretability and control, we acknowledge that its simplicity limits the impact of our findings. Though imitation by reinforcement is a data-agnostic mechanism, we have yet to explore how it behaves in more complex settings, such as using naturalistic image inputs or embodied communication. We defer to~\citet{chaabouni2022emergent,raviv_directions} for further discussion on scaling up communication settings.

A second limitation of our results is that we do not explore how imitation-based learning scales to $k > 5$ Experts. In particular, our hyperparameter regime handles up to around $k=5$ Experts-- very preliminary analyses on $k \geq 10$ Experts suggest a need to also scale up hyperparameters such as agent size and communication channel capacity. When training agents to imitate, one must therefore consider feasibility of the learning problem-- for example, as a function of the imitation network topology, communication channel size, agent size, etc-- in order for training to converge.

Finally, although our work is inspired by imitation learning in humans, the extent to which simulations explain human linguistic phenomena is not clear. We intend for our work to only serve as a testbed to understand communication from a theoretical perspective.

\section*{Ethics Statement}
Because our work uses synthetic data, it has little immediate ethical impact. However, our work may enable large populations of communicating agents down the line, which could have a range of civilian or military purposes. 

\section*{Acknowledgements}
We would like to greatly thank Marco Baroni for feedback on experiments and manuscript; Paul Michel and Rahma Chaabouni for early feedback on research direction; the three anonymous reviewers, Jeanne Bruneau-Bongard, Roberto Dessi, Victor Chomel, Lucas Weber and members of COLT UPF for comments on the manuscript. M.R. also would like to thank Olivier Pietquin, Emmanuel Dupoux and Florian Strub.

This work was funded in part by the French government under management of Agence Nationale de la Recherche as part of the ``Investissements d’avenir" program, reference ANR-19-P3IA-0001 (PRAIRIE 3IA Institute), and by the ALiEN (Autonomous Linguistic Emergence in Neural Networks) European Research Council project no.~101019291. Experiments were conducted using HPC resources from TGCC-GENCI (grant 2022-AD011013547). M.R. was supported by the MSR-Inria joint lab and granted access to the HPC resources of IDRIS under the allocation 2021-AD011012278 made by GENCI.

\bibliography{anthology,custom}
\bibliographystyle{acl_natbib}

\appendix
\renewcommand\thefigure{\thesection.\arabic{figure}}    
\renewcommand\thetable{\thesection.\arabic{table}}

\section{Expert Training}
\label{sec:expert_pretrain}
\setcounter{figure}{0}    
\setcounter{table}{0}    

\subsection{Reconstruction Task}
In the reconstruction task, a Sender observes an object with several attributes, encoding it in a message to the Receiver, and the Receiver decodes this message to reconstruct the object. Formally,
\begin{enumerate}
\item The Sender network receives a vector input $x$ and constructs a message $m$ of fixed length $L$. Each symbol is taken from the vocabulary $V = \{s_1, s_2, \cdots, s_{|V|}\}$. 
\item The Receiver network receives $m$ and outputs $\hat x$, a reconstruction of $x$. 
\item Agents are successful if $\hat x = x$.
\end{enumerate} 

\paragraph{Optimization}
In the reconstruction task, the cross-entropy loss is computed between $x$ and $\hat x$, and backpropagated directly to the Receiver. The same loss is propagated to the Sender via REINFORCE. When training with REINFORCE, we also employ an entropy regularization coefficient $\lambda$ and subtract a running mean baseline from the reward to improve training stability.

Let the Sender policy be $\pi^S$ and the Receiver be $\pi^R$. Let $x_i \in \{0, 1\}^{n_{val}}$ refer to the one-hot vector in $x$ indexed by $i$, which corresponds to one attribute. Then, the Receiver's supervised loss $\mathcal L^R$ is as follows:

\begin{align}
\mathcal L^R(m, x) &= \frac{1}{n_{att}}\sum_{i=1}^{n_{att}} CE(x_i, \pi^R(m)_i).
\end{align}

Let the Sender reward at time $t$ be $r_t = -\mathcal L^R(\pi^S(x), x)$, and let $\mu_t$ be a running mean of $r_t$. Then, the Sender's REINFORCE policy loss $\mathcal L^S$ at time $t$ is as follows:
\begin{align}
\mathcal L^S(x) = (-r_t - \mu_t) \log \pi^S(x) - \lambda \mathcal H (\pi^S(x)).
\end{align}

Finally, loss is optimized by Adam's default parameters ($\beta=0.9, 0.999$), with a learning rate of 0.005.

\subsection{Experimental Details}
We train 30 Expert pairs on the reconstruction task over 1000 epochs. Expert pairs converge to high validation accuracy and generalize to the in-distribution set well (statistics in \cref{tab:Expert_perf}).

\subsection{Expert Compositionality Distributions}
\label{sec:other_compo}
\begin{figure*}[h!]
    \centering 
    \begin{tabular}{cccc}
    \includegraphics[width=0.22\textwidth]{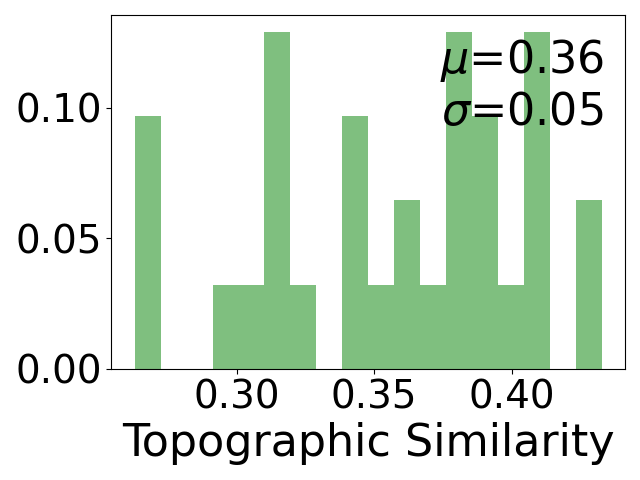} & \includegraphics[width=0.22\textwidth]{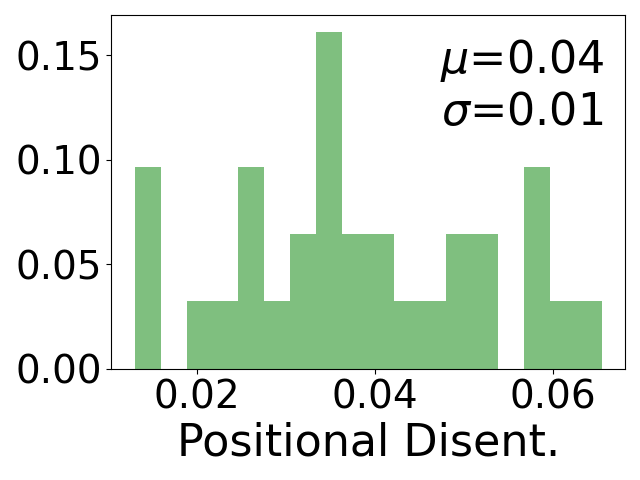} & \includegraphics[width=0.22\textwidth]{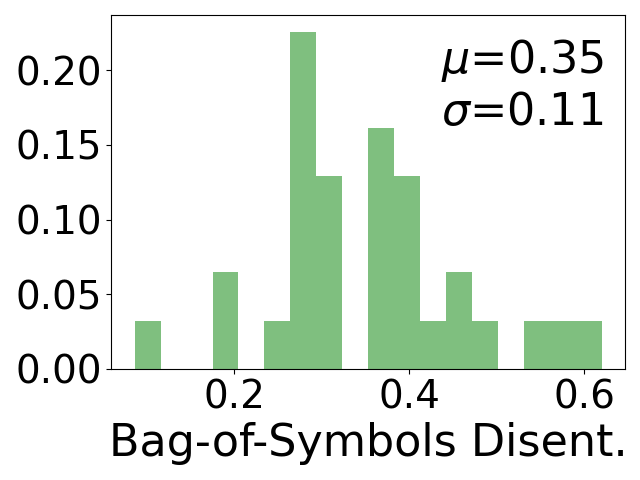} &
    \includegraphics[width=0.22\textwidth]{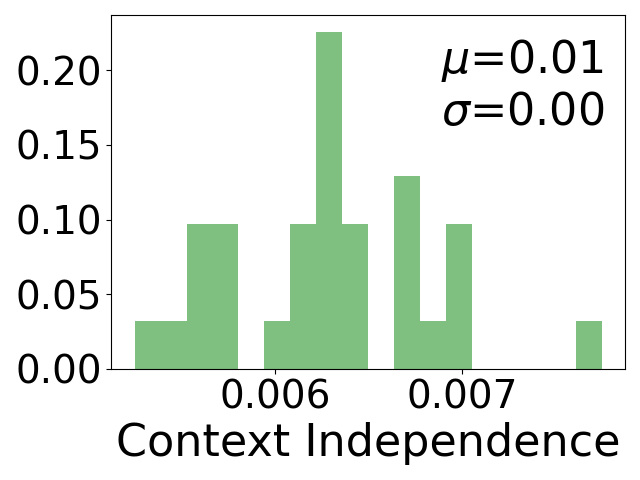}
    \end{tabular}
\caption{Distribution of 30 Expert compositionalities after 1000 epochs of training on the reconstruction task. Compositionalities are estimated on the entire validation set. According to both D'Agostino K$^2$ and Shapiro-Wilk non-normality tests, we cannot reject the null hypothesis that all four compositionality metrics follow a normal distribution ($\alpha=1e-3$).}
\label{fig:Expert_compo}
\end{figure*}

We considered using topsim, positional disentanglement (posdis)~\cite{chaabouni-etal-2020-compositionality}, bag-of-symbols disentanglement (bosdis)~\cite{chaabouni-etal-2020-compositionality}, and context independence (ci)~\cite{bogin_CI} for our experiments (see \cref{fig:Expert_compo} for distributions). However, as a fundamental reason we care about compositionality is due to its link to linguistic generalization, we focus on topsim, which we found has the highest correlation with generalization accuracy on the reconstruction task (\cref{tab:corr_compo_gen}). 

\paragraph{Topographic similarity and generalization} Similar to~\citet{rita2022on,auersperger-pecina-2022-defending} and in contrast to~\citet{chaabouni-etal-2020-compositionality,kharitonov-baroni-2020-emergent}, we find that correlations between topsim and both in-distribution and zero-shot generalization on the reconstruction task are high, and highly significant ($\alpha=1e$-$2$): Spearman's $\rho=0.83$ and Pearson's $R^2=0.81$ for in-distribution generalization, and $\rho=0.81$, $R^2=0.78$ for zero-shot generalization. This correlation is stronger than that between generalization and validation accuracy, where $\rho=0.75$ for in-distribution generalization and $\rho=0.73$ for zero-shot generalization ($\alpha=1e$-$2$). Furthermore, the correlation between topsim and validation accuracy is only $\rho=0.57$ ($\alpha=1e$-$2$) suggesting that the relationship between generalization and compositionality is not explained by high validation accuracy.\footnote{We do not report the Pearson $R^2$ for Expert validation accuracy as the its distribution violates normality assumptions according to a Shapiro-Wilk non-normality test ($\alpha=1e$-$3$)).} Our results support the stance in~\citet{auersperger-pecina-2022-defending} that compositionality, when evaluated on a suitably large dataset, indeed predicts generalization.

\begin{table}[h]
\centering
    \begin{tabular}{l|cccc}
    & \textbf{topsim} & bosdis & posdis & ci \\
    \hline 
    $\rho$ & 0.81*** & 0.74*** & 0.29 & 0.23 \\
    $R^2$ & 0.83*** & 0.78*** & 0.34* & 0.09 
    \end{tabular}
    \caption{Spearman $\rho$ and Pearson's $R^2$ correlation coefficients between compositionality metrics and in-distribution generalization accuracy on the reconstruction task.}
    \label{tab:corr_compo_gen}
\end{table}

\begin{table}[h!]
\centering
\begin{tabular}{cc}
Metric & Value \\
\hline
Validation acc. (per-object) & $0.96\pm0.03$ \\
Validation acc. (per-attribute) & $0.99 \pm 0.01$ \\
Generalization acc. (obj.) & $0.57 \pm 0.13$ \\
Generalization acc. (att.) & $0.91 \pm 0.04$ \\
Zero-shot gen. acc. (obj.) & $0.28 \pm 0.05$ \\
Zero-shot gen. acc. (att.) & $0.41 \pm 0.02$ \\
\hline
\end{tabular}
\caption{Mean and standard deviation of 30 Expert performances on the reconstruction task, first aggregated over 5 random seeds and then over the 30 Experts.}
\label{tab:Expert_perf}
\end{table}

\section{Implementation Details}
\label{sec:implementation}
\setcounter{figure}{0}    
\setcounter{table}{0}      

\subsection{Model Architecture}
\label{sec:architecture}
Both agents are single-layer recurrent neural networks that are deterministic after training. 

The Sender is a single-layer GRU~\cite{cho-etal-2014-properties} containing a fully-connected (FC) layer that maps the input $x$ to its first hidden state (dim=$128$). A symbol is then generated by applying an FC layer to its hidden state, and sampling from a categorical distribution parameterized by the output. We include LayerNorm~\cite{layernorm} after the hidden state to improve training stability. Then, the next input to the GRU is the previous output, which is sampled during training and argmax-ed during evaluation. This input is fed through an embedding module (dim=$128$), which then gets fed into the GRU. The first input is the \texttt{[SOS]} token, and the Sender is unrolled for $L=10$ timesteps to output symbols comprising a message. Only in imitation training, when unrolling the Imitator Sender, we take the Expert Sender's previous output to be the Imitator's next input so that the Imitator learns an autoregressive model of the Expert language.

The Receiver has a similar architecture to the Sender. It consists of an FC symbol-embedding layer, a GRU with LayerNorm (hidden dim=$128$), and an FC output head. The first hidden state is initialized to all zeros, then the FC-embedded symbols of the Sender message are sequentially fed into the GRU for $L=10$ timesteps. We pass the GRU's final output through a final FC layer and compute the Receiver's distribution over objects on the result, which we interpret as a concatenation of $n_{att}$ probability vectors each of length $n_{val}$.


\subsection{Hyperparameter settings}
\label{sec:hyperparams}
\setcounter{figure}{0}    
\setcounter{table}{0}    

Hyperparameters tested may be found in \cref{tab:hyperparams}. These hold for all experiments unless explicitly stated otherwise.

\paragraph{Dataset splits} Of the $n=n_{att}^{n_{val}}=10^6$ datapoints in the entire dataset, the in-distribution set has size $10^6 * (0.9)^6 = 531441$, and we randomly sample 1\% to be the training set, ($n=5315$), and delegate the rest to the generalization set ($n=526126$). Finally, the zero-shot generalization set consists of inputs where one attribute assumes the held-out value, and other attributes take on seen values ($n=354294$).

\begin{table}[h!]
\centering
\scalebox{0.8}{
\begin{tabular}{cc}
\textbf{Hyperparameter} & \textbf{Values} \\
\hline
Vocab size ($|V|$) & 10 \\
Message length ($L$) & 10 \\
\# Attributes ($n_{att}$) & 6 \\
\# Values ($n_{val}$) & 10 \\
\hline 
Learning rate & 0.005 \\
Batch size & 1024 \\
Entropy coeff. ($\lambda$) & 0, 0.01, 0.1, 0.5, 1 \\
\hline 
GRU hidden size & 128 \\
GRU embedding size & 128 \\
\hline 
\hline 
Expert pretraining epochs & 1000 \\
Single imitation training epochs (RF) & 2000 \\
Single imitation training epochs (SV) & 500 \\
\# Experts in imitation mixture ($k$) & 2--5 \\
Sender imitation training epochs & 2000 \\
Rcvr imitation training epochs & 7000 \\
\hline
\end{tabular}}
\caption{From top to bottom: the communication channel, optimization, architectural, and experimental hyperparameters, respectively. All hyperparameters pertain to both Sender and Receiver unless otherwise stated. The number of training epochs (bottom section) is selected based on visual inspection of convergence of validation accuracy curves.}
\label{tab:hyperparams}
\end{table}

\subsection{Implementation Details}
Experiments were implemented using PyTorch and the EGG toolkit~\cite{kharitonov:etal:2021}. They were carried out on a high-performance cluster equipped with NVIDIA GPUs. The number of GPU-hours to run all experiments is estimated to be between 50 and 100.

\section{Supplementary Material: Ease-of-Imitation}
\label{sec:appendix}
\setcounter{figure}{0}    
\setcounter{table}{0}      
In the compositionality vs. ease-of-imitation experiments, we train newly initialized Imitator pairs on each Expert pair over 500 epochs for supervision and 2000 epochs for reinforcement, aggregating over 3 random seeds. The number of training epochs is chosen by visual inspection of validation accuracy convergence. We note that, when imitating by both reinforcement and supervision, there is no initial increase in topsim followed by a convergence to Expert topsim (\cref{fig:no_topsim_hump}), contrary to what is observed in~\cite{Ren2020Compositional}.

For imitation by reinforcement, we use an entropy coefficient of $\lambda=0.1$ for both Sender and Receiver. Comparing \texttt{SOL} and $T$ for both Sender and Receiver to other compositionality metrics (\cref{tab:spearman_bc}), we see that topsim is generally most correlated with sample complexity and speed-of-learning. For the opposite reason, we did not move ahead with, e.g., experiments on positional disentanglement.

\begin{figure}
    \includegraphics[width=0.45\textwidth]{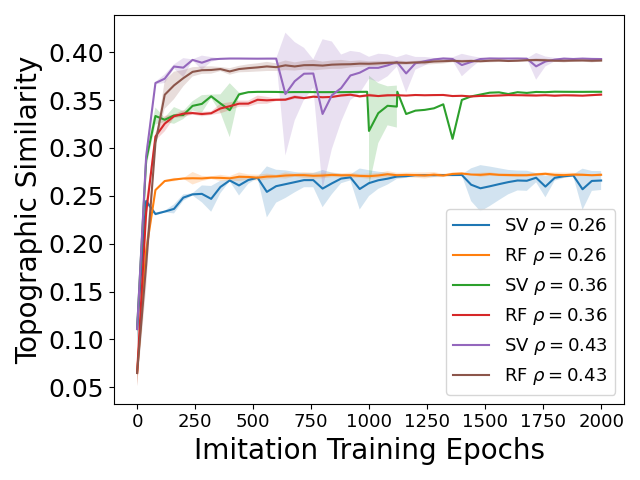}
    \caption{Evolution of topsim of an Imitator Sender when being trained separately via REINFORCE (RF) and supervision (SV) on Expert topsims ($\rho$) of 0.26, 0.36, 0.43, corresponding to low, average, and high values in the distribution of Expert topsims.}
    \label{fig:no_topsim_hump}
\end{figure}

\begin{figure*}[h!]
    \centering 
    \begin{tabular}{cc}
    \includegraphics[width=0.45\textwidth]{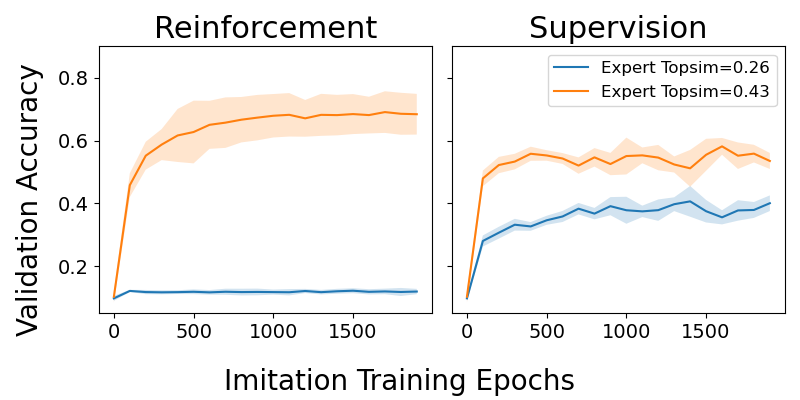} & \includegraphics[width=0.45\textwidth] {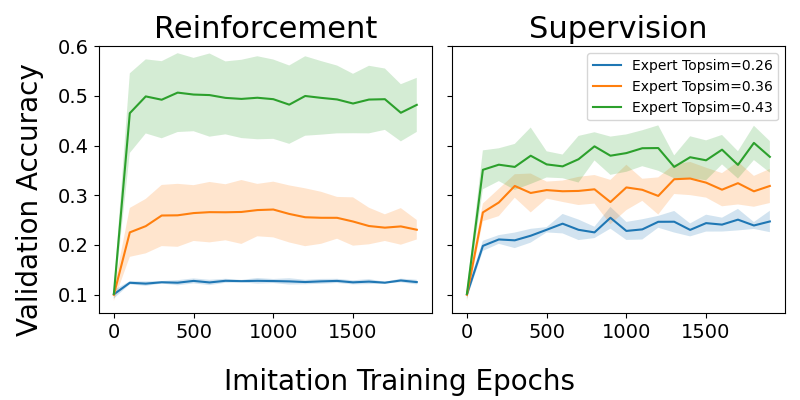} \\
    \includegraphics[width=0.45\textwidth]{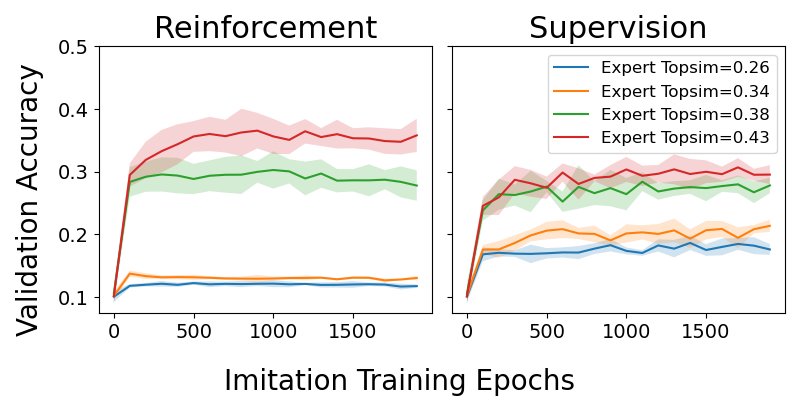} &
    \includegraphics[width=0.45\textwidth]{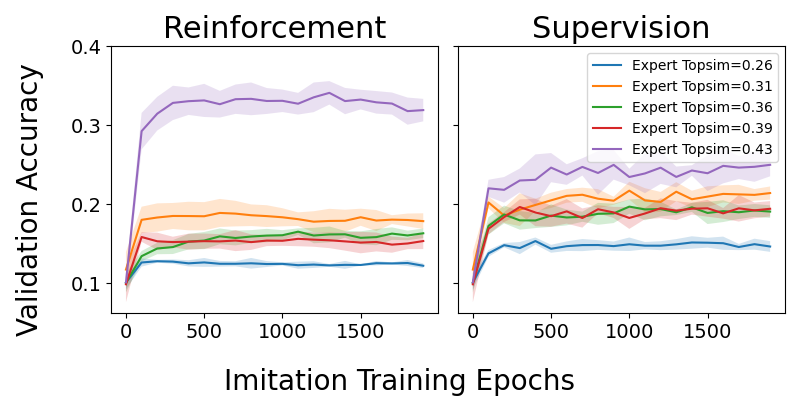}
    \end{tabular}
\caption{For $k=2$ (top left), $k=3$ (top right), $k=4$ (bottom left) and $k=5$ (bottom right) Experts, validation accuracy curves of the Imitator Sender on Expert messages over 2000 epochs of imitation training. For each setting of $k$, reinforcement is shown on the left and supervision on the right. Note (1) the greater dispersion of validation accuracy when training by reinforcement compared to be supervision; (2) higher ``selection", or validation accuracy, on the best topsim Expert in reinforcement compared to supervision; (3) lower validation accuracy on the poorest topsim Expert in RF compared to SV. 
}
\label{fig:imi_val}
\end{figure*}


\begin{table*}[h]
    \centering
    \scalebox{0.9}{
    \begin{tabular}{rl|l l | l l | l l | l l}
    & & \multicolumn{2}{c}{$T_S$} & \multicolumn{2}{c}{$T_R$} & \multicolumn{2}{c}{\texttt{SOL}$_S^I$} & \multicolumn{2}{c}{\texttt{SOL}$_R^I$} \\
    \hline 
     &  & $\rho$ & $R^2$ & $\rho$ & $R^2$ & $\rho$ & $R^2$ & $\rho$ & $R^2$ \\
    \hline
    \textbf{SV} & topsim & -0.65*** & -0.66*** & -0.80*** & -0.80*** & 0.65 *** & 0.65*** & 0.75*** & 0.76*** \\
    & bosdis & -0.64*** & -0.67*** & -0.54*** & -0.60*** & 0.63*** & 0.71*** & 0.81*** & 0.83*** \\
    & CI & -0.24 & -0.16 & -0.20 & -0.01 & 0.40** & 0.34* & 0.30* & 0.09 \\
     & posdis & -0.15 & -0.18 & -0.26 & -0.26 & 0.22 & 0.23 & 0.17 & 0.16 \\
     & \emph{acc} & -0.53*** & -- & -0.72*** & -- & 0.56 *** & -- & 0.53*** & -- \\
      \hline
    \textbf{RF} & topsim & -0.66*** & -0.66*** & -0.60*** & -0.68*** & 0.45*** & 0.41** & 0.59*** & 0.63*** \\
    & bosdis & -0.73*** & -0.72*** & -0.61*** & -0.67*** & 0.71*** & 0.75*** & 0.41** & 0.40** \\
    & CI & -0.24 & -0.16 & -0.41** & -0.39** & 0.11 & -0.06 & 0.32* & 0.29 \\
     & posdis & -0.03 & -0.11 & -0.43** & -0.38** & -0.1 & -0.15 & 0.25 & 0.26 \\
     & \emph{acc} & -0.51*** & -- & -0.52*** & -- & 0.28 & -- & 0.23 & -- \\
     \hline
    \end{tabular}}
    \caption{Spearman $\rho$ and Pearson $R^2$ correlations between compositionality of Expert communication and learning speed of Imitators for imitation by supervision (\textbf{SV}) and reinforcement (\textbf{RF}). For comparison, correlations between validation accuracy of Expert communication (acc) and learning speed are also reported. Correlations are shown with significance determined by a two-sided Pearson's R significance test ($\alpha = 0.01$ (***), $0.05$ (**), $0.1$ (*)).}
    \label{tab:spearman_bc}
\end{table*}

\section{Supplementary: Imitators Select Compositional Languages to Learn}
\label{sec:app_selection}
\setcounter{figure}{0}    
\setcounter{table}{0}   

\begin{figure}
\includegraphics[width=0.48\textwidth]{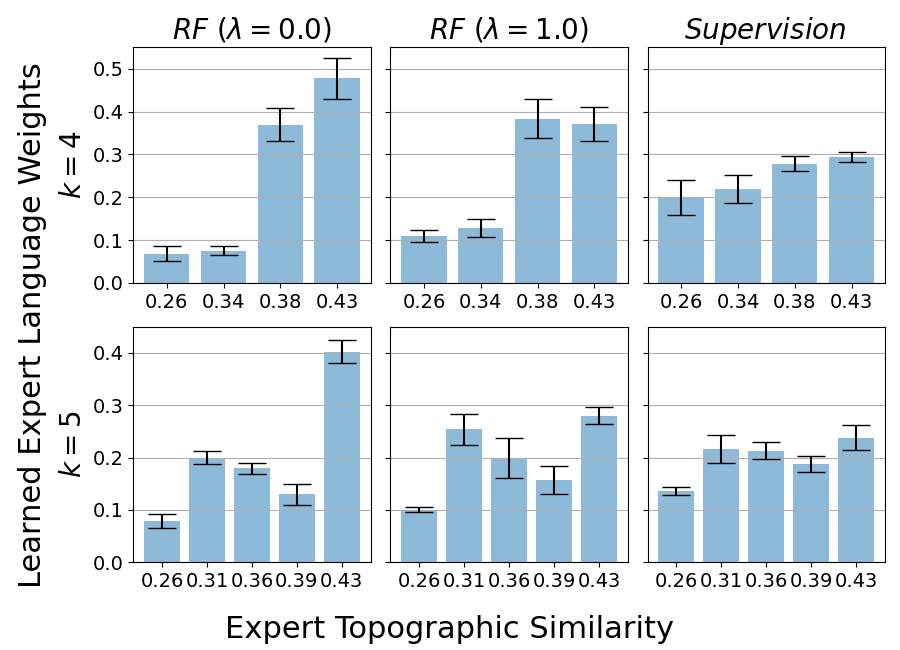}
\caption{Imitator's learned weights on $k=4$ (top row) and $k=5$ (bottom row) languages, where the topsim of Expert languages ranges uniformly from 0.26 to 0.43. From left to middle, as the entropy regularization coefficient $\lambda$ increases, the distribution appears more uniform. At the limit, the weight distribution appears most uniform when the Imitator is trained by supervision (right column). Refer to \cref{fig:skew_entropy_4_5} for the skews and entropies of the distributions.}
\end{figure}

\begin{figure*}
\begin{tabular}{cc}
\includegraphics[width=0.47\textwidth]{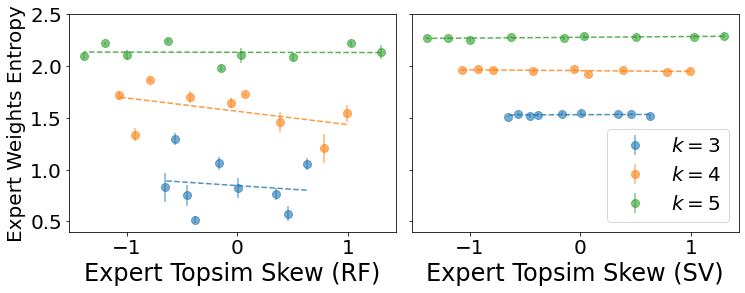} &
\includegraphics[width=0.47\textwidth]{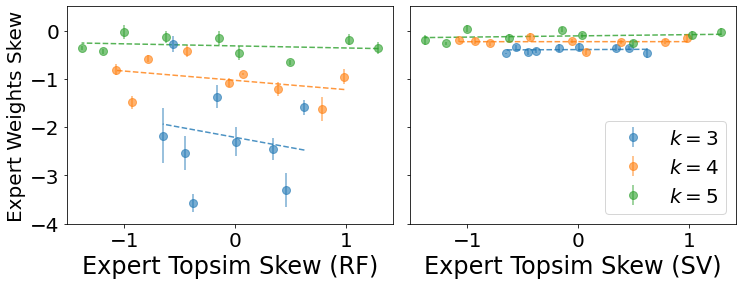} \\
(a) & (b) 
    \end{tabular}
    \caption{In Sender imitation, the shape of the learned Expert weights is independent of the shape of the Expert topsim skew for both reinforcement (RF) and supervision (SV). Shape of the Expert weight distribution is quantified by [Entropy, Skew], and we plot [Entropy, Skew] x Skew for the learned Expert weights against the Expert topsim skew for $k = 3, 4, 5$, and for both RF and SV (omitting $k=2$ because skew is artificially $0$-- the plot would look like a vertical line). Robustness is seen in the lack of a significant positive or negative trend in the data for all numbers of Experts tested.}
    \label{fig:robust_to_skew_entropy_composite}
\end{figure*}

\begin{figure}
\includegraphics[width=0.47\textwidth]{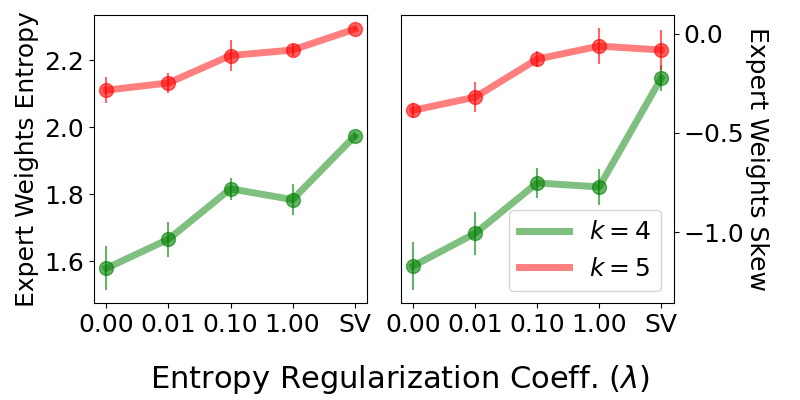}
\caption{The entropy (left) and skew (right) ($\pm 1$ standard deviation) of learned Expert weights by a Sender Imitator for $k=4$ and $5$ Experts. Values attain maxima in the supervision setting (SV).}
\label{fig:skew_entropy_4_5}
\end{figure}

\subsection{Optimization}
In the imitation task, we test both direct supervision and REINFORCE. Importantly, when doing a forward pass for the Sender during training, we feed it the Expert symbol from the previous timestep as input so that the Sender learns an autoregressive model of language. Hence, define the Imitator policy $\pi_j^I$ as in \cref{sec:appendix_proof}.

In the direct supervision setting, for the Sender producing a distribution over messages $\pi^I$ given $x$ and a given Expert $i$ producing message $m^{(i)}$, where $m_j$ is the $j^{th}$ symbol of $m^{(i)}$, the overall loss for a uniform mixture of $k$ Expert Senders is the following cross-entropy loss:

\begin{align}
    \mathcal L^{I}_{SV}(x) = \sum_{i=1}^k \sum_{j=1}^{L} CE \left(m^{(i)}_j, \pi^{I}_j\right).
\end{align}

 In the REINFORCE setting, we use accuracy per-symbol as a reward for the Sender, with entropy regularization and mean baseline.\footnote{We also tried REINFORCE using negative cross-entropy loss as a reward, but found training to be unstable.} For Expert $i$, this corresponds to a reward $r^{(i)}$ of

 \begin{align}
r^{(i)} = \frac{1}{L} \sum_{j=1}^{L} Acc \left(m^{(i)}_j, \pi^{I}_j\right)
 \end{align}
 
 and a policy loss of

 \begin{align}
    \mathcal L^{I, (i)}_{RF}(x) = (- r^{(i)} - \mu_t) \log \pi^I(x) - \lambda \mathcal H(\pi^I(x)),
\end{align}

per Expert, which is averaged over Experts to produce the mixture-policy loss. This is optimized by Adam with a learning rate of $0.005$.

\subsection{Sampling Sender Expert Distributions}
\label{sec:shape}

To test the effect of the shape (skew, standard deviation) of the Expert topsims on imitation, we define a set of 10 distributions for each setting of $k>2$ Experts, noting that when $k=2$, the skew is mechanically equal to 0.

For interpretability, we hold the endpoints of the distributions equal at the minimum and maximum possible topsims (0.26, 0.43) for all distributions and values of $k$. Then, we sample the median $\mathcal M$ of the 10 distributions evenly from 0.26 to 0.43. Then, we fill the other $k-3$ points to create a uniform distribution with mean $\mathcal M$. If the median is less than the average topsim, then the left endpoint of this uniform distribution is the minimum topsim. Otherwise, it is the maximum topsim. 

\subsection{Effect of population size on learned Expert Sender weights}
We find that the selection effect decreases as the number of Experts increases, i.e. the Expert weight distribution looks increasingly uniform (\cref{fig:robust_to_skew_entropy_composite}). We offer two possible explanations: (1) the harder learnability of this problem given our hyperparameter regime-- this is suggested by the lower maximum validation accuracy achieved on any one Expert in RF-- or (2) the (mechanically) smaller variance between values, holding endpoints equal, as we increase the number of agents. Notably, the purpose of this work is not to scale up to imitation in large populations of agents; we delegate the problem of operationalizing RL imitation at scale to future work.

\subsection{Learning a uniform mixture of policies}
\label{sec:appendix_proof}
\paragraph{Claim} A Sender that imitates a uniform mixture of $k$ Expert Senders will output a uniform mixture of the $k$ Expert languages.

\paragraph{Proof} Let $\pi^{(1)}\cdots \pi^{(k)}$ be $k$ Expert Senders and let $\pi^I$ be the Imitator Sender. For each position in a message, agents produce a probability distribution over all possible symbols in $V$. Recall that the Expert Senders are deterministic at evaluation time. Given an input $x$, we write $m^{(i)}_j$
as the value of the $j^{\text{th}}$ position in the message $m^{(i)}$ produced by Expert $i$. 

For the Imitator Sender, we write
$$\pi^I_j := \pi^{I}\left(m^{(i)}_{j-1} \ ; \ x\right) \in [0,1]^{|V|}$$
as the probability distribution over possible symbols in position $j$ of a message produced by the Imitator agent, given the previous output symbol $m_{j-1}^{(i)}$ of Expert $i$. The $k^{\text{th}}$ index of $\pi_j^I$, or $\pi_j^I[k]$, gives the Imitator agent's probability of symbol $k$ at position $j$ in the message.

The ideal Imitator $\pi^{I*}$ minimizes the cross-entropy objective between its messages and that of a uniform mixture of $k$ Expert Senders. Formally,
\begin{align*}
    \pi^{I*} &= \min_{\pi^I} \sum_{i=1}^k \sum_{j=1}^{L} CE\left(m^{(i)}_j, \pi^{I}_j\right) \\
    &= \min_{\pi^I} \sum_{i=1}^k \sum_{j=1}^{L} -\log \pi^I_j[m_j^{(i)}] \\
    &= \max_{\pi^I} \sum_{i=1}^k \sum_{j=1}^{L}\log \pi^I_j[m^{(i)}_j] \\
    &= \max_{\pi^I} \prod_{i=1}^k \prod_{j=1}^{L}\pi^I_j[m^{(i)}_j] \ \ \text{subj. to } \sum_{k\in V} \pi_j^I[k] = 1.
\end{align*}
whose unique solution is $\pi^I_j[m^{(i)}_j] = \pi^I_j[m^{(l)}_j]$ $\forall j \in \mathbb{N}_{L}$, $\forall i \neq l \in \mathbb{N}_k$, i.e. a uniform distribution over Expert languages. $\Box$

\section{Receiver imitation}
\label{sec:receiver_imitation}
\setcounter{figure}{0}    
\setcounter{table}{0}    
\subsection{Setup}

The Receiver imitation task is as follows: given a set of $k$ Expert Receivers and their corresponding Senders, we train an identical, newly initialized Receiver on the Experts' inputs and outputs $(m, \hat x)$. That is, for each round of training, all $k$ Experts as well as the Imitator Receiver receive input $m$, or the output of Expert Sender given $x$, and output $\hat x^{(1)} \cdots \hat x^{(k)}$ and $\hat x^I$, respectively. Imitators are then tasked to minimize the difference between their output and a uniform mixture of the $k$ Expert outputs.

The architecture for the Receiver agent may be found in \ref{sec:architecture}.

\paragraph{Optimization} Similar to in Sender imitation, we test a supervised learning and a reinforcement imitation learning setting. For supervised learning, the Receiver imitation loss is equal to the cross-entropy loss between its output and the Expert Receiver's output given the same corresponding Expert Sender's message $m$. Then, the loss over the entire mixture is the average cross-entropy loss per Receiver, aggregated across Expert Receivers.

For REINFORCE, the Receiver reward is similar to the Sender reward-- analogous to the per-symbol accuracy, it is the per-attribute accuracy. We compute the corresponding policy loss (using a mean baseline per Expert and $\lambda$ defined in \cref{tab:hyperparams}), and average over all Experts to get the overall policy loss for the Receiver.

\subsection{Imitation and Selection of Compositionality}
With the large communication channel size typical of emergent communication games, we can expect little Expert message collision. Then, in this setting, Receiver imitation consists of learning a many-to-one mapping of messages to outputs, obviating a real need for selection if the goal is to maximize eventual communication accuracy. Indeed, we find that Imitator Receivers learn to be multilingual, achieving high validation accuracy on all Experts, and especially in the supervised setting. 

\begin{figure}[h!]
    \includegraphics[width=0.45\textwidth]{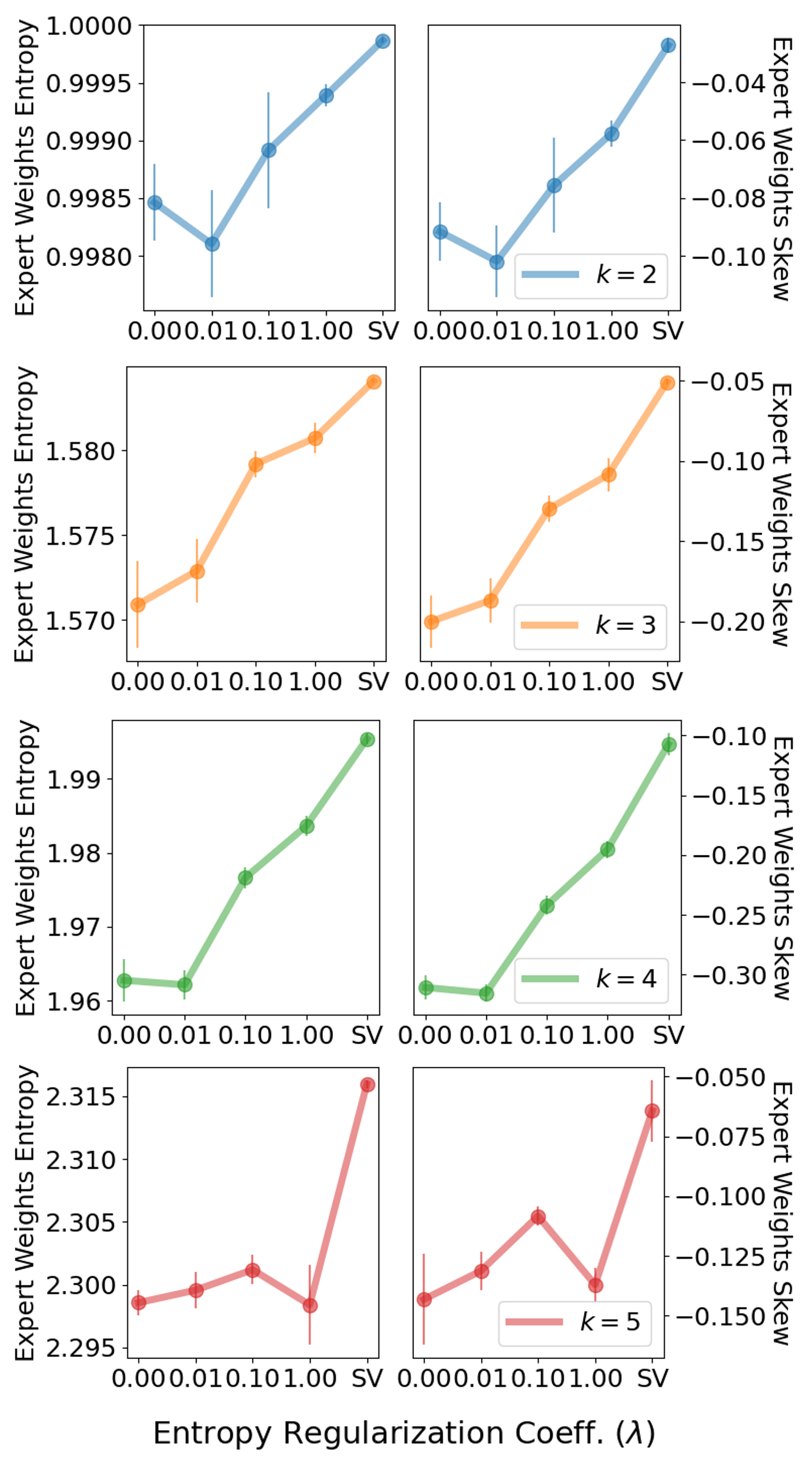}
    \caption{For the Receiver imitation, the entropy (left) and skew (right) ($\pm 1$ standard deviation) of learned Expert weights for $k=2$--$5$ Experts. We plot one row per Expert for legibility as the entropy ranges are quite different for different values of $k$. Values generally increase as exploration ($\lambda$) increases, attaining maxima in the supervision setting (SV).}
    \label{fig:receiver_imitation_skew_entropy}
\end{figure}

\begin{figure}[h!]
    \includegraphics[width=0.47\textwidth]{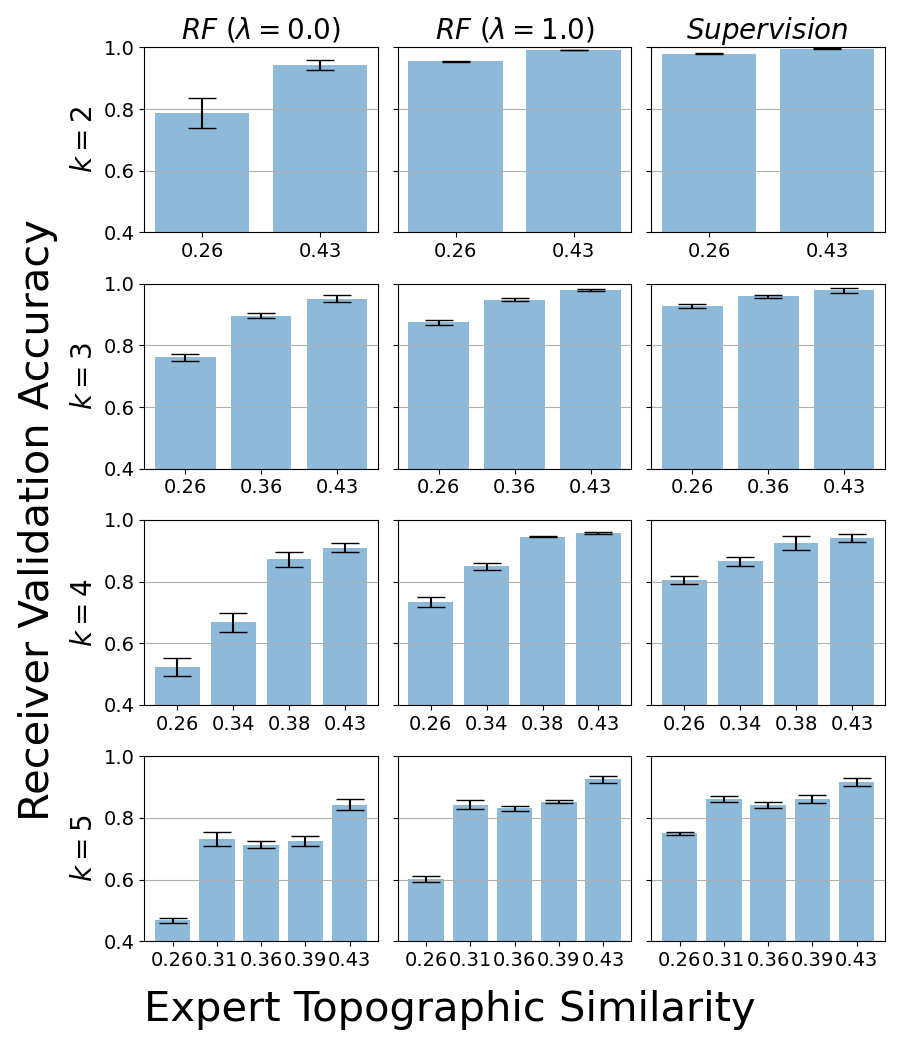}
    \caption{Receiver Imitator's validation accuracies per-Expert on $k=2$-$5$ (top to bottom) languages, where the topsim of Expert languages ranges evenly from 0.26 to 0.43. Imitation by supervision results in the highest and most uniform validation accuracies, and increasing the entropy coefficient $\lambda$ in imitation by reinforcement increases the uniformity of validation accuracies. The $y$-axis is cut below at 0.4 for legibility.}
    \label{fig:receiver_imitation}
\end{figure}

\begin{figure*}[h!]
    \centering 
    \begin{tabular}{cc}
    \includegraphics[width=0.45\textwidth]{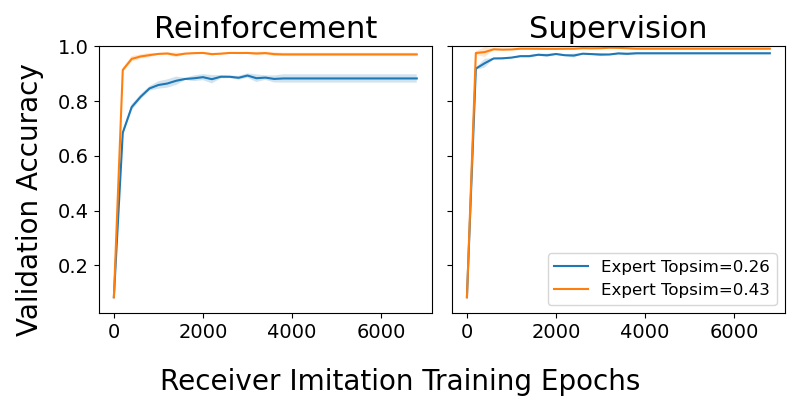} & \includegraphics[width=0.45\textwidth] {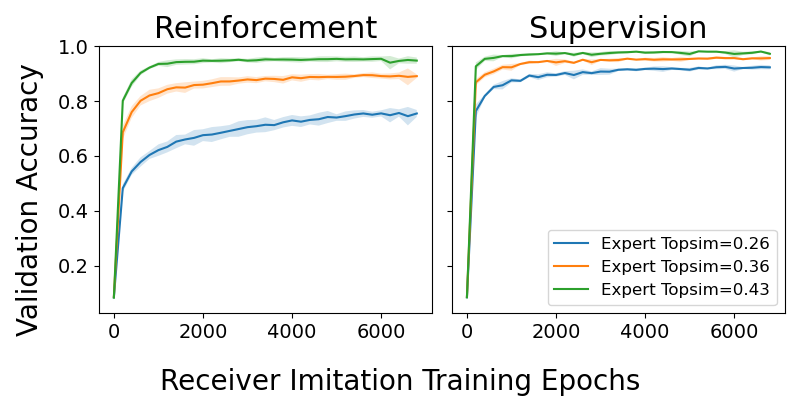} \\
    \includegraphics[width=0.45\textwidth]{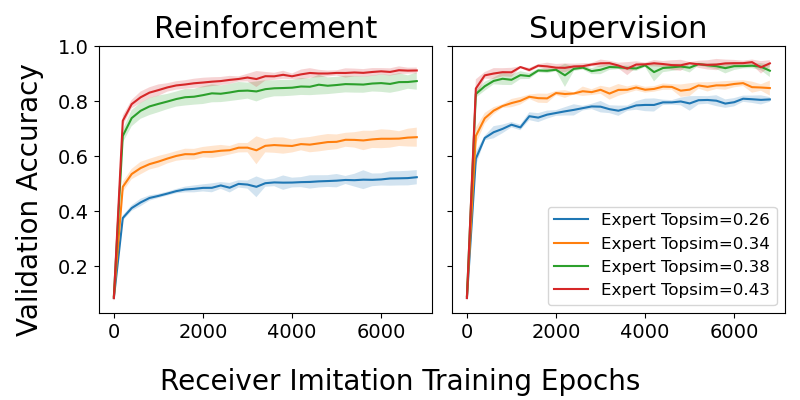} &
    \includegraphics[width=0.45\textwidth]{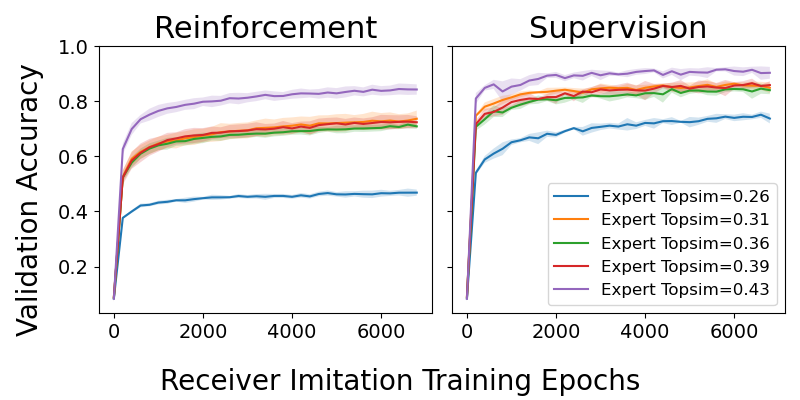}
    \end{tabular}
\caption{For $k=2$ (top left), $k=3$ (top right), $k=4$ (bottom left) and $k=5$ (bottom right) Experts, validation accuracy curves of the Imitator Receiver on Expert messages over 7000 epochs of imitation training. For each setting of $k$, reinforcement is shown on the left and supervision on the right. Note (1) the greater dispersion of validation accuracy when training by reinforcement compared to be supervision; (2) lower validation accuracy on the poorest topsim Expert in RF compared to SV.}
\label{fig:receiver_imi_val}
\end{figure*}

We do note, however, greater differentiation in validation accuracy, as well as speed-of-learning, between Experts of varying compositionality when using reinforcement compared to supervision, and again influenced by the entropy coefficient $\lambda$ (\cref{fig:receiver_imitation_skew_entropy,fig:receiver_imitation,fig:receiver_imi_val}). 

How one operationalizes Receiver imitation then depends on one's goal: for example, if the goal is to maximize communication accuracy in a population of communicating agents, then we want to have ``tolerant" Receivers, and imitation by supervision allows the Receiver to achieve the highest validation accuracy on all languages. However, if we want to bottleneck the compositionality of the language in the population, we want to have more ``selective" Receivers, and imitation by reinforcement may be more appropriate.

\subsection{Speed-of-Imitation May Explain Compositional Selection}
Results for the Sender also hold for the Receiver; see \cref{sec:sol} for the analogous comments.

\end{document}